\documentclass[letterpaper]{article} % DO NOT CHANGE THIS
\usepackage{aaai25}  % DO NOT CHANGE THIS
\usepackage{times}  % DO NOT CHANGE THIS
\usepackage{helvet}  % DO NOT CHANGE THIS
\usepackage{courier}  % DO NOT CHANGE THIS
\usepackage[hyphens]{url}  % DO NOT CHANGE THIS
\usepackage{graphicx} % DO NOT CHANGE THIS
\usepackage{natbib}  % DO NOT CHANGE THIS AND DO NOT ADD ANY OPTIONS TO IT
\usepackage{caption} % DO NOT CHANGE THIS AND DO NOT ADD ANY OPTIONS TO IT
\usepackage{algorithm}
\usepackage{algorithmic}
\usepackage{amsmath}
\usepackage{booktabs}
\usepackage{bbding}
\usepackage{amssymb}
\usepackage{xcolor}

\usepackage{newfloat}
\usepackage{listings}
\DeclareCaptionStyle{ruled}{labelfont=normalfont,labelsep=colon,strut=off} % DO NOT CHANGE THIS
\floatstyle{ruled}
\newfloat{listing}{tb}{lst}{}
\floatname{listing}{Listing}
\title{Topo2Seq: Enhanced Topology Reasoning via Topology Sequence Learning}
\author{
    Yiming Yang\textsuperscript{\rm 1,\rm 2}, Yueru Luo\textsuperscript{\rm 1,\rm 2}, Bingkun He\textsuperscript{\rm 3}, Erlong Li\textsuperscript{\rm 4}, Zhipeng Cao\textsuperscript{\rm 4},\\ Chao Zheng\textsuperscript{\rm 4}, Shuqi Mei\textsuperscript{\rm 4}, Zhen Li\textsuperscript{\rm 2,\rm 1}
    \thanks{Corresponding author.}
}
\affiliations{
    \textsuperscript{\rm 1}FNii-Shenzhen, Shenzhen, China\\
    \textsuperscript{\rm 2}SSE, CUHK-Shenzhen, Shenzhen, China\\
    \textsuperscript{\rm 3}SCSE, Wuhan University, Wuhan, China\\
    \textsuperscript{\rm 4}T Lab, Tencent, Beijing, China\\
    \{yimingyang@link., lizhen@\}cuhk.edu.cn
}

\newcommand{\modelname}{Topo2Seq }
\usepackage{bibentry}
\begin{document}

\maketitle

\begin{abstract}
Extracting lane topology from perspective views (PV) is crucial for planning and control in autonomous driving. This approach extracts potential drivable trajectories for self-driving vehicles without relying on high-definition (HD) maps. However, the unordered nature and weak long-range perception of the DETR-like framework can result in misaligned segment endpoints and limited topological prediction capabilities. Inspired by the learning of contextual relationships in language models, the connectivity relations in roads can be characterized as explicit topology sequences. In this paper, we introduce Topo2Seq, a novel approach for enhancing topology reasoning via topology sequences learning. The core concept of Topo2Seq is a randomized order prompt-to-sequence learning between lane segment decoder and topology sequence decoder. The dual-decoder branches simultaneously learn the lane topology sequences extracted from the Directed Acyclic Graph (DAG) and the lane graph containing geometric information. Randomized order prompt-to-sequence learning extracts unordered key points from the lane graph predicted by the lane segment decoder, which are then fed into the prompt design of the topology sequence decoder to reconstruct an ordered and complete lane graph. In this way, the lane segment decoder learns powerful long-range perception and accurate topological reasoning from the topology sequence decoder. Notably, topology sequence decoder is only introduced during training and does not affect the inference efficiency. Experimental evaluations on the OpenLane-V2 dataset demonstrate the state-of-the-art performance of Topo2Seq in topology reasoning.
\end{abstract}


\section{Introduction}
In recent years, lane topology reasoning in autonomous driving has gained increasing attention \cite{li2023lanesegnet,li2024enhancing,ma2024roadpainter}. This is because autonomous driving has traditionally relied on offline HD maps to provide path information. However, road conditions can be uncertain and challenging, and outdated offline HD maps can be disastrous for autonomous vehicles \cite{liao2023lane}. Solely relying on these maps is insufficient to meet the higher demands of advanced autonomous driving, such as L4 and L5 levels.

\begin{figure}[h]
    \centering
    \includegraphics[width=0.88\linewidth]{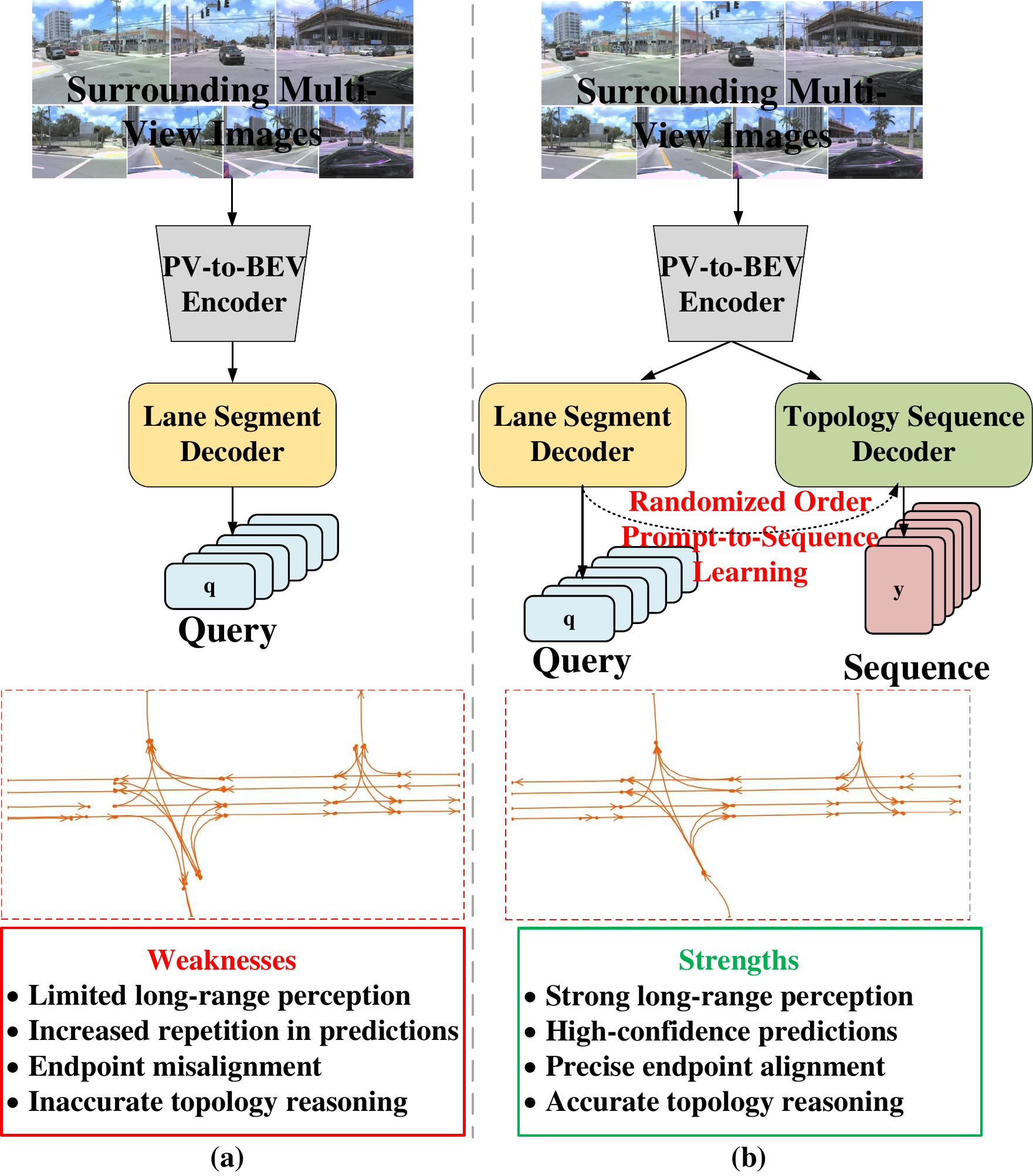}\\
    %\vspace{-0.5em}
    \caption{Comparison between previous methods (a) and~\modelname (b). Due to the limited sampling positions for each query in deformable-DETR and the unordered detection characteristics, existing methods exhibit several weaknesses. (b)~\modelname employs
    %randomized order 
    a prompt-to-sequence learning strategy, which enhances lane segment perception and topology reasoning through topology sequence learning.}
\label{fig:motivation}       % Give a unique label

\end{figure}

To address these issues, autonomous vehicles need to perform lane topology reasoning, which involves perceiving the surrounding road flow in real-time from panoramic images and extracting both the geometric positions of the centerlines and their topological relationships. Therefore, lane topology reasoning is crucial for trajectory prediction and planning in end-to-end autonomous driving \cite{can2021structured,wu2023topomlp,li2023graph}.

Early works focused on the characterization of map elements. The semantic map learning approaches assign labels to each pixel to mark map elements \cite{li2022hdmapnet,liu2023petrv2,liu2023vectormapnet}. However, inconsistent semantic predictions and insufficient instance perception can cause confusion in the planning of autonomous driving systems. Moreover, pixel-by-pixel post-processing often fails in topology extraction and causes time-consuming problems \cite{liao2023lane}. As an upgraded alternative, vectorized map learning approaches employ learnable queries to extract map elements by vectorized lines and polygons in DETR-like framework. These methods offer high detection accuracy and fast inference speeds \cite{liu2023vectormapnet,liao2022maptr}. Nevertheless, they lack comprehensive trajectory detection in map learning, such as lane direction and connectivity. To address these issues, recent studies on lane topology reasoning have transformed the centerline topology into lane graphs \cite{li2023lanesegnet,li2023graph}. These end-to-end networks are designed to predict both the line segments, characterized by ordered points, and the topological relationships, represented by an adjacency matrix. However, these methods do not explicitly model the relationships between lanes, instead relying on MLPs to determine the connection probabilities between queries. Due to weak long-range perception and unordered detection characteristics in DETR-like framework \cite{carion2020end}, simple MLPs struggles to effectively learn the connectivity between lanes. As a result, existing methods meet several weaknesses as illustrated from Fig.\ref{fig:motivation} (a). In language models, sequence learning can capture contextual relationships in long texts while maintaining the correct order \cite{vaswani2017attention,devlin2018bert}. Inspired by language models, representing lane graph as sequences can explicitly capture the geometric positions and topological relationships of lane. However, in sequence-to-sequence approaches \cite{peng2024lanegraph2seq,lu2023translating}, the auto-regressive model depends on prior predictions to generate subsequent outputs, resulting in considerable inefficiencies (approximately 0.1 FPS) \cite{lu2023translating} due to the need for repeated inference.

In this paper, we introduce $\textbf{Topo2Seq}$, a novel approach that enhances topology reasoning via topology sequence learning. Topo2Seq utilizes a dual-decoder architecture, comprising a lane segment decoder and a topology sequence decoder. The topology sequence decoder predicts lane topology sequences extracted from a Directed Acyclic Graph (DAG), while the lane segment decoder extracts lane graphs containing geometric information. Randomized order prompt-to-sequence learning is then employed to extract unordered key points from the lane graph predicted by the lane segment decoder. These key points are input into the prompt design of the topology sequence decoder, enabling the reconstruction of an ordered and complete lane graph. In this way, the lane segment decoder gains powerful long-range perception and accurate topological reasoning from the topology sequence decoder through a shared encoder as illustrated from Fig.\ref{fig:motivation} (b). Notably, the topology sequence decoder is only introduced during training and does not impact inference efficiency.

The contributions of this paper can be summarized as follows:
\begin{itemize}
    \item We present Topo2Seq, a novel framework with dual-decoder training for enhancing topology reasoning by leveraging topology sequence learning.
    \item We explicitly model lane graph as sequences to capture long-range geometric positions and topological relationships of lanes.
    \item We introduce a randomized order prompt-to-sequence learning mechanism, which enables the lane segment decoder to gain robust long-range perception and accurate topology reasoning capabilities from the topology sequence decoder.
    \item Extensive experiments conducted on the multi-view topology reasoning benchmark OpenLane-V2 \cite{wang2024openlane} demonstrate the state-of-the-art performance of Topo2Seq in topology reasoning.

\end{itemize}

\begin{figure*}[h]
	\centering
	\includegraphics[width=1.0\linewidth]{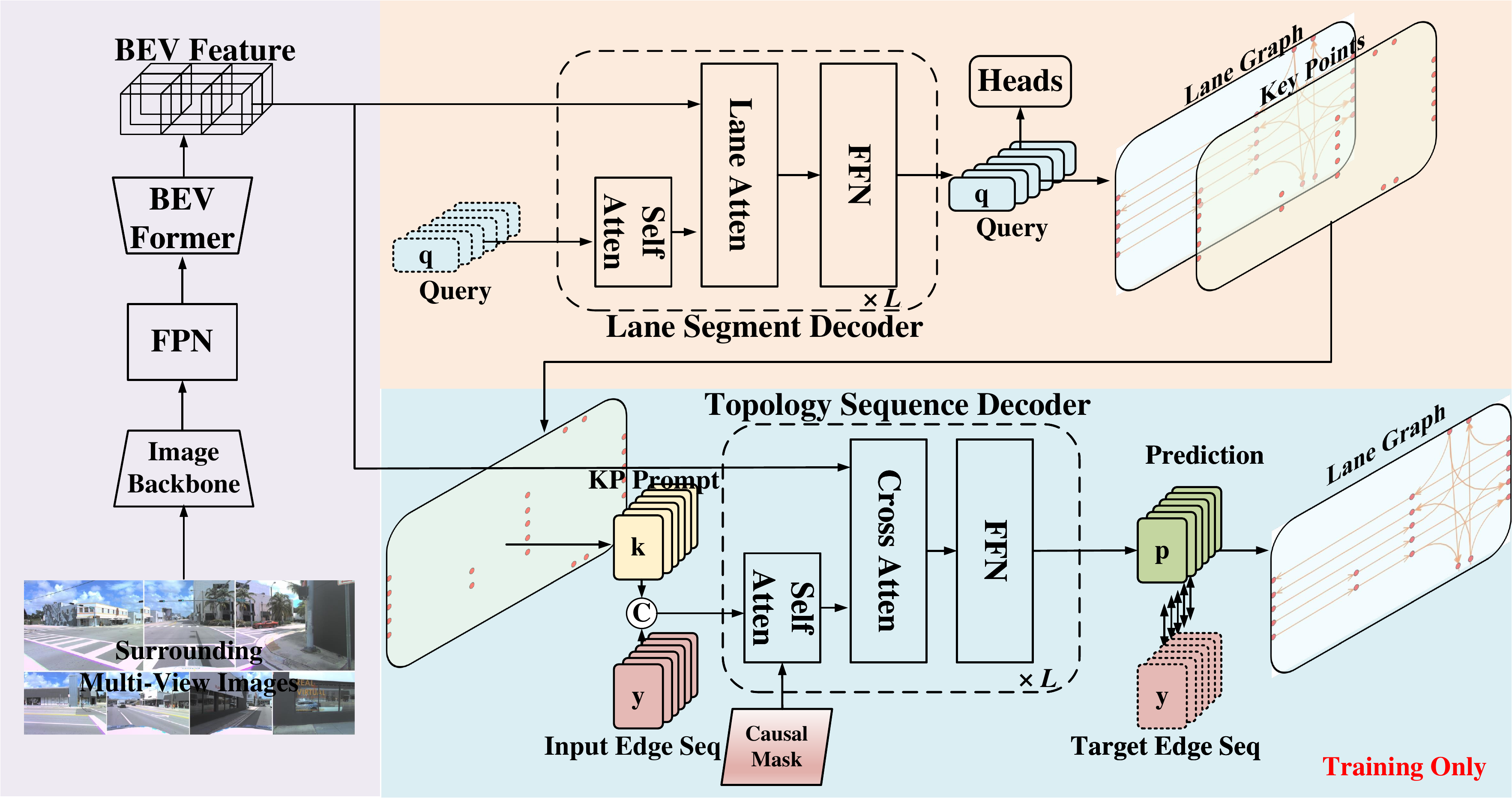}\\
	\caption{The framework of Topo2Seq. Topo2Seq is composed of three main components. First, the surrounding multi-view images are processed by the image backbone, FPN, and BEVFormer to generate bird's-eye view (BEV) features. The lane segment decoder then predicts the lane graph. The key points from this predicted lane graph are fed into the topology sequence decoder to construct key point prompts, which are subsequently concatenated with edge sequences. The topology sequence decoder infers the relationships between discrete key points and reconstructs them into a coherent lane graph. By doing so, the topology sequence decoder enhances the BEV features with improved long-range dependencies and contextual integration, thereby aiding the lane segment decoder in topology reasoning.}
	\label{fig:framework}       % Give a unique label
\end{figure*}

\section{Related Work}
\textbf{Online Map Learning}
Recent advancements in online map learning focus on detecting map elements using onboard sensors to construct local high-definition (HD) maps. Traditionally, this task has been approached as a pixel-level semantic segmentation problem \cite{liu2023petrv2,liu2023vision}. To mitigate the time-consuming post-processing and shape ambiguity issues \cite{li2022hdmapnet}, recent approaches have shifted towards learning vectorized representations of map elements. For instance, VectorMapNet \cite{liu2023vectormapnet} introduces a vectorized HD map learning framework that predicts a sparse set of polylines from a bird's-eye view. The MapTR series \cite{liao2022maptr,liao2023maptrv2} leverage hierarchical query embedding to more effectively learn geometrical shapes at both point and shape levels. To further enhance positional embedding in queries, MapQR \cite{liu2024leveraging} introduces a scatter-and-gather query mechanism that encodes shared content across different positions. However, current online map learning methods still fall short in providing detailed lane information, such as lane direction and topological structure. While these methods are effective at detecting map elements, they are not equipped to deliver the detailed trajectories needed for planning in end-to-end autonomous driving.

\textbf{Topology Reasoning}
Topology Reasoning primarily focuses on centerline perception and connectivity relations. STSU \cite{can2021structured} is the first end-to-end framework to detect centerlines and objects using learnable queries from a BEV perspective. The centerline queries are processed by detection, control, and association heads to generate a lane graph, which is widely followed in most subsequent works. TopoNet \cite{li2023graph} represents lane connectivity as a lane graph and designs a scene graph neural network to refine the position and shape of lanes. LaneSegNet \cite{li2023lanesegnet} introduces lane attention and identical initialization to enhance long-range perception. RoadPainter \cite{ma2024roadpainter} generates centerline masks to better refine points with large curvature. To fully leverage higher recall 2D results, Topo2D \cite{li2024enhancing} updates 3D lane queries using 2D lane priors. TopoMLP \cite{wu2023topomlp} enhances topology results by incorporating lane point coordinates as positional embeddings. Most of these methods are built on a DETR-like framework \cite{zhu2020deformable}. However, the unordered nature and weak long-range perception of the DETR-like framework limit topology reasoning. In this paper, we address these challenges by introducing sequence-to-sequence learning.

\textbf{Visual Sequence-to-Sequence Learning}
Visual sequence-to-sequence learning relies on pixel-based observations, transforming downstream task objectives into a language-like format, i.e., sequences. Pix2Seq \cite{chen2021pix2seq} quantizes and serializes detected objects into sequences of discrete tokens, and by dequantizing the output sequences from the auto-regressive transformer, the detected bounding boxes are obtained. Pix2Seqv2 \cite{chen2022unified} introduces task-specific prompts at the beginning of sequences, enabling training for multiple vision tasks, such as detection, segmentation, keypoint detection, and captioning. However, the aforementioned tasks do not emphasize the order and relationships between instances, which are crucial for topology reasoning. To enable lane graph extraction in a sequence-to-sequence manner, RoadNet \cite{lu2023translating} integrates landmarks, curves, and topology into a unified sequence representation. LaneGraph2Seq \cite{peng2024lanegraph2seq} transforms lane graphs into a combination of vertex and edge sequences. Nevertheless, sequence-to-sequence learning, which relies on auto-regressive transformers, tends to be slow in inference. To fully leverage the strengths of sequence-to-sequence learning for long-range modeling and relationship extraction, we incorporate it into the training phase to enhance feature extraction, rather than using it directly for lane graph inference.

\section{Method}
\subsection{Problem Formulation}
Given multi-view images $\mathcal{I}$ captured by a vehicle's surround-view cameras, the goal of Topo2Seq is to perceive centerlines and their topology. Centerlines are represented as a list of ordered 3D points $L=\left [ k^0,\cdots,k^{n-1}\right ]$, where n is set to 10 and each 3D point is denoted as $k_i = (x,y,z) \in \mathbb{R} ^3$. The topology is described by an adjacency matrix $A\in \mathbb{R} ^{m \times m}$, where $m$ is the number of detected centerlines. In this matrix, $A_{ij}=1$ means the endpoint of lane $L_i$ coincides with the starting point of lane $L_j$. However, it is generally challenging to achieve precise alignment of the starting and end points of connected lane predicted by the network.

\subsection{Overview}
As depicted in Fig. \ref{fig:framework}, Topo2Seq takes multi-view images as input. The image backbone \cite{he2016deep}, FPN \cite{lin2017feature}, and BEVFormer \cite{li2022bevformer} are utilized to encode these multi-view images into a BEV feature $\mathcal{F} \in \mathbb{R} ^{H\times W \times C}$. The lane segment decoder updates learnable query $Q^l$ through self-attention and cross-attention with the BEV feature $\mathcal{F}$. The updated queries are passed to prediction heads for lane segmentation. The key points from the starting and end points of predicted lane graph are extracted as key point prompts $y^K$. Ground Truth (GT) lane graph is translated into edge sequences $y^E$. Edge sequences are concatenated with key point prompts $y^K$ and are fed into topology sequence decoder, where they interact with BEV features $\mathcal{F}$. The prediction sequences are supervised by the target edge sequences and can be dequantized into a lane graph. During testing, the topology sequence decoder does not perform inference. Instead, the lane segment decoder predicts the centerlines and the adjacency matrix.

\textbf{Lane segment decoder}\quad We denote a set of instance-level queries as $\{Q^l_j\}^{N_L}_{j=1}$, where $N_L$ is the preset number of queries, which is usually greater than the number of centerlines in the lane graph. The queries are feed into lane segment decoder to obtain the updated queries:
\begin{equation}
	\label{equation1}
	\hat{Q}^l  = \textbf{LaneDec}(\mathcal{F},Q^l)
\end{equation}
where \textbf{LaneDec} denotes Lane segment decoder. Within each lane segment decoder layer, the lane queries are sequentially updated through a self-attention module, a lane attention module \cite{li2023lanesegnet}, and a feed-forward network.

\textbf{Heads}\quad We emplopy MLPs to generate 3D coordinates of lane $L$ and topology $A$. The topology between lanes are predicted by:
\begin{equation}
	\label{equation2}
	\hat{Q}^l_{emb_1}, \hat{Q}^l_{emb_2} = MLP(\hat{Q}^l),MLP(\hat{Q}^l)
\end{equation}
\begin{equation}
	\label{equation3}
	A = Sigmod(MLP(Concat(\hat{Q}^l_{emb_1},\hat{Q}^l_{emb_2})))
\end{equation}
where the MLPs are independent of each other. To provide a more detailed representation of the lane graph, we can predict not only the topology but also the offsets of the left and right lane boundaries, the types of these boundaries, and pedestrian crosswalks.

\textbf{Topology sequence decoder}\quad We follow \cite{chen2021pix2seq} to build topology sequence decoder. Each decoder layer includes a self-attention module, a cross attention module, and a feed-forward network. Auto-regressive property is maintained by causal mask in self-attention module. The whole structure brings several advantages in extracting and refining the BEV features: (1) \textbf{Enhanced feature refinement:} The model can selectively focus on relevant areas of the BEV features based on sequence. This targeted attention helps refine BEV features by emphasizing regions critical for accurately reconstructing the lane graph or understanding the scene. (2) \textbf{Improved long-range dependencies:}The topology sequence decoder enhances the capture of long-range dependencies between distant key points. This injects the contextual information between key points into the lane segment decoder, enabling it to predict more aligned lane segment endpoints. (3) \textbf{Contextual integration:} By focusing on specific key point prompts, the model can reduce the impact of irrelevant or redundant information in the BEV features. This results in more efficient feature extraction and potentially reduces noise in the final predictions. The output from training the topology sequence decoder can be represented as:

\begin{equation}
	\label{equation1}
	\hat{y}^E  = \textbf{TopoSeqDec}(\mathcal{F},Concat(y^K,y^E))
\end{equation}
where \textbf{TopoSeqDec} represents the topology sequence decoder, and $\hat{y}^E$ denotes the predicted edge sequence.

\textbf{Structure of the sequences}\quad Following RoadNet \cite{lu2023translating}, we transform the Directed Acyclic Graph (DAG) into edge sequences. Each key point in a lane serves as either a starting point or an endpoint, and each edge can be represented by six integers:
\begin{equation}
	\label{equation4}
	y^E=[int(x_i), int(y_i), cls, con, int(bx_i), int(by_i)]
\end{equation}
where the first two integers $int(x_i), int(y_i)$ represent the discretized coordinates of the key point. $cls$ indicates the category of key point, which can be Ancestor, Lineal, Offshoot, or Clone. The $con$ denotes the connection of the key point. If $cls$ is either Ancestor or Lineal, then $con$ is set to 0. Otherwise, $con$ is set to the index of the parent key point. Since a cubic Bezier curve can effectively represent the trajectory of a lane between two key points, the last two integers $int(bx_i), int(by_i)$ indicate the intermediate control points of the Bezier curve. To determine a unique order of key points, we select the location at the right front in the BEV perspective as the starting point and use Depth-First Search to perform the sorting.

\begin{figure}[t]
	\centering
	\includegraphics[width=1.0\linewidth]{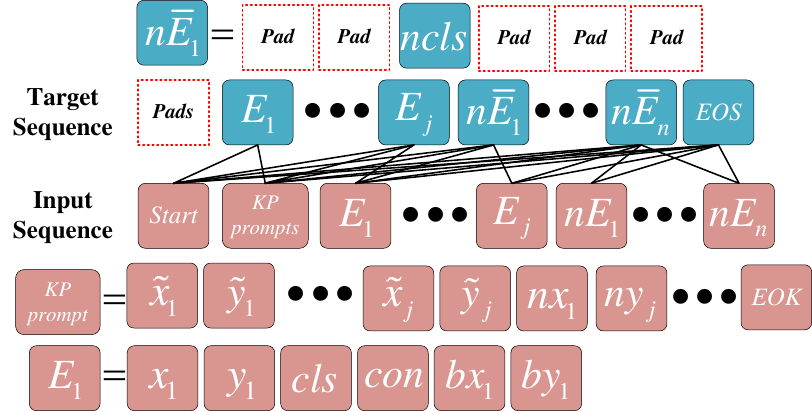}\\

	\caption{The illustration of input and target sequences.}
	\label{fig:sequences_modeling}       % Give a unique label
\end{figure}

\begin{table*}[h]
\centering
\begin{tabular}{c|cc|cccc}
\hline
Method & Backbone & Epochs & mAP $\uparrow $ & $\text{AP}_{ls} \uparrow $ & $\text{AP}_{ped} \uparrow$ & $\text{TOP}_{lsls} \uparrow$ \\ \hline\hline
  TopoNet \cite{li2023graph} & ResNet-50 &24&  23.0   &   23.9     &   22.0      &     -      \\
  MapTR \cite{liao2022maptr}  & ResNet-50 &24 &   27.0  &     25.9   &    28.1     &   -        \\
   MapTRv2 \cite{liao2023maptrv2} & ResNet-50 &24  &   28.5  &  26.6      &   30.4      &     -      \\ 
   LaneSegNet \cite{li2023lanesegnet}  & ResNet-50  &24 &   33.4 &  31.9      &   34.9      &     25.4      \\ 
  LaneSegNet \cite{li2023lanesegnet}  & ResNet-50  &48 &   36.4  &  34.9      &   37.9      &     27.3      \\\hline \hline
     \textbf{Topo2Seq (ours)}   & ResNet-50 &24 &    33.6  &   33.7      &    33.5      &      26.9      \\

   \textbf{Topo2Seq (ours)}   & ResNet-50 &48  &    \textbf{37.7}  &   \textbf{36.9}      &    \textbf{38.5}      &      \textbf{29.9}      \\
   \hline
\end{tabular}
\caption{Comparison with the state-of-the-arts on OpenLane-V2 benchmark on lane segment. mAP ($\%$), $\text{AP}_{ls}$ ($\%$), $\text{AP}_{ped}$ ($\%$), and $\text{TOP}_{lsls}$ ($\%$) are reported. }\label{tab:comparison_lane_segment}
\end{table*}

During training, we construct two types of sequences as illustrated in Fig. \ref{fig:sequences_modeling}: the input sequence and the target sequence used for supervision. The input sequence starts with the $<Start>$ token, followed by the key point prompts, then the GT edges $y^E$, and the remaining length is filled with noise edges $y^{nE}$ \cite{chen2021pix2seq}. Key point prompts $y^K$ include key points for all predicted edges as well as noise edges. Notably, the key points for all predicted edges are unordered and do not correspond to the order of the coordinates in the edge sequence. Finally, the key point prompts conclude with the $<EOK>$ token. In the target sequence, the positions of key point prompts are filled with $<pad>$ tokens, followed by the ground truth edges and noise edges, and ending with $<EOS>$. To help the topology sequence decoder identify which edges are noise edges, the supervised noise edges are marked with the noise class $<ncls>$ at their category positions, while other positions are filled with $<pad>$ tokens. 
The $<pad>$ tokens are excluded from loss calculation.

\textbf{Randomized order prompt-to-sequence learning}\quad 
The lanes predicted by the lane segment decoder often have misaligned endpoints. Representing two lanes requires four endpoints, which may exhibit geometric inconsistencies between the endpoints of different lanes. In contrast, the edge sequence uses only three points to represent two adjacent lane lines with perfectly aligned endpoints, improving trajectory comprehension for autonomous driving. To leverage the long-range understanding and sequential relationship capabilities of sequence-to-sequence learning, we facilitate interaction between the lane segment decoder and the sequence topology decoder at the key point prompts.

Based on the predictions from the lane segment decoder, we rank the predicted lanes by confidence from highest to lowest and filter out any duplicate key points in each predicted lane using predicted adjacency matrix:
\begin{equation}
	\label{equation5}
y^K\gets \begin{cases}
 k^{n-1}_j &\quad A_{ij}=1\\k^{0}_j,k^{n-1}_j &\quad A_{ij}=0

\end{cases}
\end{equation}
where the coordinates of key points are discretized. 
The object of randomized order prompt-to-sequence learning can be expressed as:
\begin{equation}
	\label{equation6}
max\sum_{i=1}^{L}w_ilogP(\hat{y}^E_{i}|,Concat(y^K,y^E_{< i}),\mathcal{F} ) 
\end{equation}
where $w_i$ denotes the class weight, y$^E_{< i}$ indicates all tokens before y$^E_i$, and $\hat{y}^E$ is predicted target sequence. The input key point prompts are unordered relative to the edge sequence, enabling the sequence topology decoder to guide the network in inferring relationships between discrete key points. In this way, the network infers the correct associations among unordered key point prompts, compelling it to focus on long-range relationships. Additionally, this process encourages the network to refine the positions of high-confidence key points, reducing duplicate predictions and aligning endpoints in BEV domain.
By enhancing interaction between two decoders, this approach indirectly addresses the limitations in capturing long-range relationships via sequence-to-sequence learning.

\textbf{Loss function}\quad The overall loss function in Topo2Seq is defined as follows:
\begin{equation}
	\label{equation7}
\mathcal{L} = \alpha_1\mathcal{L}_{1} + \alpha_2\mathcal{L}_{cls} + \alpha_3\mathcal{L}_{seg} + \alpha_4\mathcal{L}_{lt} + \alpha_5\mathcal{L}_{top} + \alpha_6\mathcal{L}_{seq}
\nonumber
\end{equation}
Where $\mathcal{L}_{1}$ represents a L1 loss. $\mathcal{L}_{cls}$ denotes a focal loss \cite{lin2017focal} for lane classification. $\mathcal{L}_{seg}$ includes a cross-entropy loss and a dice loss. $\mathcal{L}_{lt}$ represents a cross-entropy loss for classifying the left and right lane types (e.g., non-visible, solid, dashed). $\mathcal{L}_{top}$ is a focal loss used to supervise the relationship information between the predicted adjacency matrix $A$ and the GT adjacency matrix $\tilde{A}$. $\mathcal{L}_{seq}$ indicates a maximum likelihood loss that supervises the topology sequence decoder in predicting tokens. The weights for each loss are denoted by $\alpha_1$, $\alpha_2$, $\alpha_3$, $\alpha_4$, $\alpha_5$, and $\alpha_6$.

\begin{table*}[h]

\centering
\begin{tabular}{c|cc|ccc}
\hline
Method  & Backbone & Epochs& $ \text{OLS}^*\uparrow  $ & $\text{DET}_{l} \uparrow  $ &  $\text{TOP}_{ll} \uparrow$ \\ \hline\hline
  VectorMapNet \cite{liu2023vectormapnet} & ResNet-50 &24&13.8 &  11.1  &     2.7          \\
       STSU \cite{can2021structured}  & ResNet-50  &24  &    14.9   &12.7 &   2.9        \\
  MapTR  \cite{liao2022maptr}  & ResNet-50 &24 &21.0 &   17.7  &     5.9      \\
     TopoNet \cite{li2023graph} & ResNet-50 &24 &30.8 &   28.6      &     10.9      \\

   Topo2D \cite{li2024enhancing}  & ResNet-50 &24  &38.2 &   29.1      &     26.2      \\
     TopoMLP \cite{wu2023topomlp}   & ResNet-50  &24 &37.4 &   28.3      &     21.7     \\
      TopoMLP$^*$ \cite{wu2023topomlp}  & Swin-B  &48 &33.5&   32.5      &     11.9     \\
       RoadPainter$^*$ \cite{ma2024roadpainter}  & ResNet-50 &24  &29.4 &   30.7      &     7.9      \\
   LaneSegNet \cite{li2023lanesegnet}  & ResNet-50 &24  &40.7 &   31.1      &     25.3      \\
      LaneSegNet \cite{li2023lanesegnet}  & ResNet-50 &48  &43.3 &   34.3      &     27.3      \\\hline \hline
     \textbf{Topo2Seq (ours)}   & ResNet-50 &24 &    42.7  &  33.5     &    27.0 \\

   \textbf{Topo2Seq (ours)}   & ResNet-50 &48 &    \textbf{45.8}  &   \textbf{36.7}      &    \textbf{30.0}     \\
   \hline
\end{tabular}
\caption{Comparison with the state-of-the-arts on OpenLane-V2 benchmark on centerline perception. $\text{OLS}^*$ ($\%$), $\text{DET}_{l}$ ($\%$), $\text{TOP}_{ll}$ ($\%$), and $\text{TOP}_{lsls}$ ($\%$) are reported. The $\text{OLS}^*$ is calculated between $\text{DET}_{l}$ and $\text{TOP}_{ll}$. The centerlines from LaneSegNet and Topo2Seq are extracted from the lane segment results. $*$ denotes metrics from the $\text{TOP}_{ll}$ v 1.0.0 version extracted from the referenced paper.}\label{tab:comparison_centerline}
\end{table*}

\begin{table*}[h]
\centering
\begin{tabular}{c|ccc|cccc}
\hline
Index  & OP & RP & RPL  & mAP $\uparrow $ & $\text{AP}_{ls} \uparrow $ & $\text{AP}_{ped} \uparrow$ & $\text{TOP}_{lsls} \uparrow$  \\ \hline

   1     &  \Checkmark &    &    &  32.5   &   31.5     &   33.5      &   25.1        \\
   2     &   &  \Checkmark  &    & 34.9    &  33.4      &   36.3      &   27.8         \\ 
  3     &   &   &  \Checkmark   & 33.8    &  32.3      &   35.4     &   26.5      \\ 
   4    &   &  \Checkmark  &   \Checkmark &   37.7  &     36.9   &   38.5      &      29.9  \\ \hline
\end{tabular}
\caption{Ablation study on the OpenLane-V2 benchmark: OP, RP, and RPL refer to the ordered GT key points prompts, randomized order GT key points prompts, and randomized order prompt-to-sequence learning, respectively. }\label{tab:ablation}
\end{table*}

\section{Experiments}
\subsection{Dataset}
We evaluate our Topo2Seq model on the OpenLane-V2 dataset \cite{wang2024openlane}, a recently released open-source dataset specifically designed to focus on topology reasoning in autonomous driving. OpenLane-V2 is derived from Argoverse2 \cite{wilson2023argoverse} and nuScenes \cite{caesar2020nuscenes} datasets. The data spans various global locations and includes challenging scenarios such as daytime and nighttime, sunny and rainy conditions, as well as urban and suburban environments. In this paper, we primarily evaluate Topo2Seq on the $\mathit{subset_A}$, which consists of 7 surrounding images per frame. The training set includes approximately 27,000 frames, and the validation set contains around 4,800 frames.

\subsection{Implementation Details}
We employ ResNet-50 pre-trained on ImageNet \cite{deng2009imagenet} as our image backbone. FPN \cite{lin2017feature} is used to obtain multi-scale features. The BEVFormer \cite{li2022bevformer} encodes the multi-scale features of the surround view image into BEV features. Consistent with recent works, the BEV perception range is set to cover the x-axis from $[-50.0m, +50.0m]$ and the y-axis from $[-25.0m, +25.0m]$. The BEV grid is configured to be $200\times100$. 
The configuration of the Lane Segment Decoder follows that of LaneSegNet. It consists of 6 layers with the number of queries set to 200. The model achieves a frame rate of 14.7 FPS. Both the input and target sequences consist of 802 tokens, comprising a 201-token key point prompt and a 601-token edge sequence. Due to resource limitations, we train our network on 4 NVIDIA A100 GPUs with a total batch size of 4. We ensure that each sample underwent the same number of iterations with recent works. The initial learning rate is $2\times10^{-4}$ with a cosine annealing schedule during training. AdamW \cite{diederik2015adam} is adopted as optimizer. The values of $\alpha_1$, $\alpha_2$, $\alpha_3$, $\alpha_4$, $\alpha_5$, and $\alpha_6$ are set to 0.025, 1.5, 3.0, 0.1, 5.0, and 1.0, respectively. We evaluate three training strategies: (1) Training the network for 24 epochs to achieve stable output using randomly ordered GT key points as key point prompts, followed by 24 epochs focusing on the interaction between the two decoders. (2) Conducting 24 epochs of training solely on the interaction between the two decoders. (3) Training for 12 epochs to achieve stable output using randomly ordered GT key points as key point prompts, followed by 12 epochs of decoder interaction.

\subsection{Metrics}
We evaluate Topo2Seq on two types of tasks: lane segment perception and centerline perception. For lane segment perception, we use mAP, $\text{AP}_{ls}$, $\text{AP}_{ped}$, and $\text{TOP}_{lsls}$. For centerline perception, we use $\text{DET}_{l}$, $\text{TOP}_{ll}$, and a redefined OLS between $\text{DET}_{l}$ and $\text{TOP}_{ll}$. It is noted that in the OpenLane-V2 benchmark, centerline and lane segment labels are misaligned, and lane segment labels are more challenging to detect due to their higher number of lane pieces.

\begin{figure*}[h]
	\centering
	\includegraphics[width=0.92\linewidth]{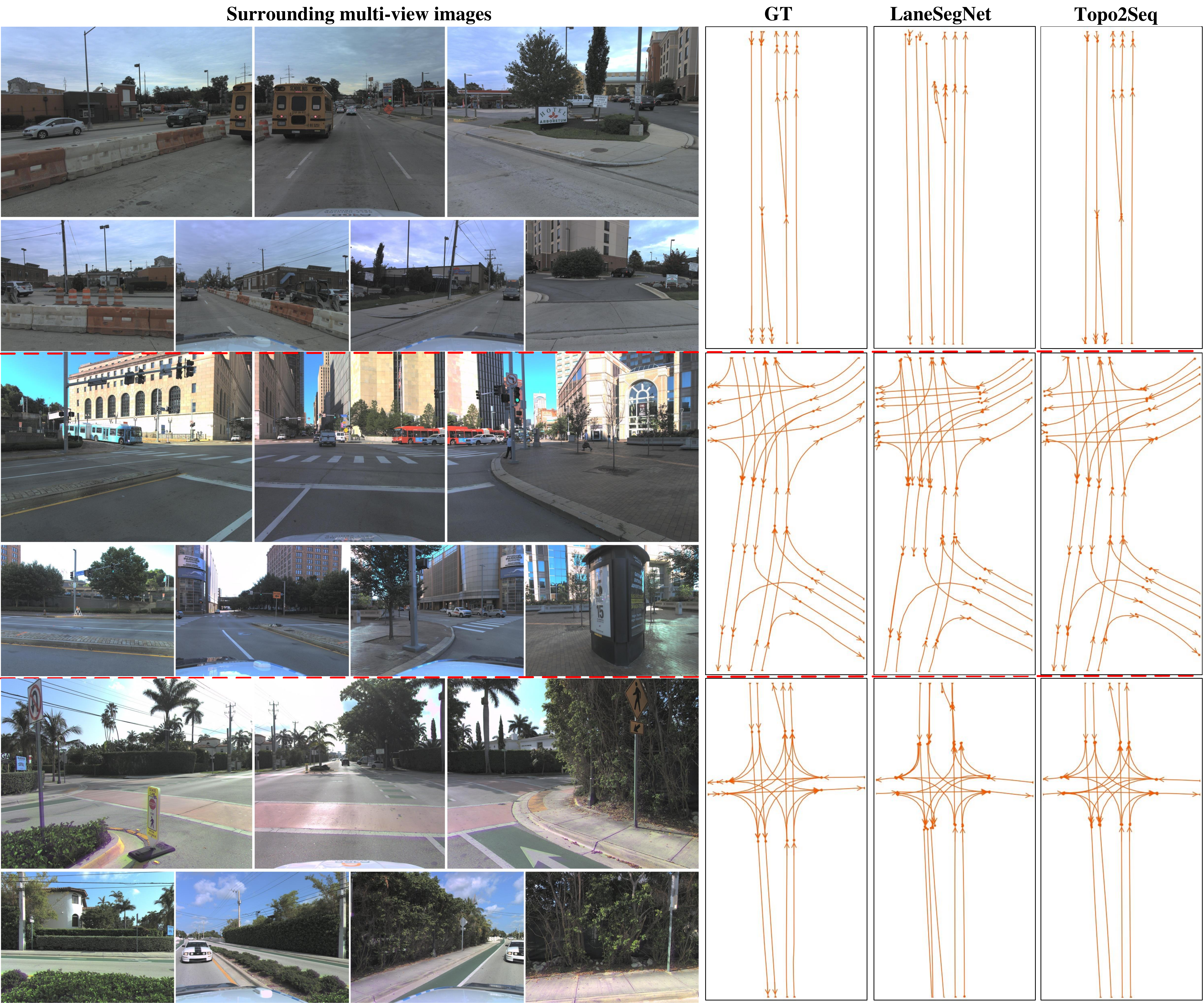}\\
	%\vspace{-0.5em}
	\caption{Qualitative results. Compared to LaneSegNet, Topo2Seq, with the assistance of sequence learning, produces higher-quality lane graphs.}
	\label{fig:Qualitative_results}       % Give a unique label
\end{figure*}

\subsection{Main Results}
We first compare our Topo2Seq with state-of-the-art methods on OpenLane-V2 benchmark on lane segment. The results on the OpenLane-V2 $\mathit{subset_A}$ are shown in Tab. \ref{tab:comparison_lane_segment}. The results of the state-of-the-art methods are obtained primarily from their respective papers. When trained for 24 epochs (12 epochs for achieving stable output followed by 12 epochs of decoder interaction), Topo2Seq outperforms LaneSegNet by $1.8\%$ in AP$_{ls}$ and $1.5\%$ in TOP$_{lsls}$. With a two-stage training process over a total of 48 epochs using ResNet-50, Topo2Seq achieves a $37.7\%$ mAP and $29.9\%$ TOP$_{lsls}$. Under the same configuration, Topo2Seq surpasses LaneSegNet by $2.0\%$ in AP$_{ls}$ and $2.6\%$ in TOP$_{lsls}$.

The results of centerline perception on the OpenLane-V2 $\mathit{subset_A}$ are displayed in Tab. \ref{tab:comparison_centerline}. With the same 24 epochs of training, Topo2Seq outperforms LaneSegNet by $2.0\%$ in OLS$^*$, $2.4\%$ in DET$_l$, and $1.7\%$ in TOP$_{ll}$. Compared with TopoMLP and LaneSegNet with the same 48 training epochs, Topo2Seq exceeds TopoMLP by $4.2\%$ improvement in DET$_{l}$ and performs better than Lanesegnet by $2.5\%$ in OLS$^*$, $2.4\%$ in DET$_{l}$, and $2.7\%$ in TOP$_{ll}$. These results indicate that introducing an additional sequence decoder interaction during training allows the network to achieve considerable improvements in topology reasoning.

\subsection{Ablation Studies}
We have studied each important design in Topo2Seq. The ablation studies are shown in Tab. \ref{tab:ablation}. When introducing ordered GT key point prompts into sequence learning, the network is only capable of learning the trajectories between key points, without being forced to infer the relationships between them. This explains why the results in index 2 outperform those in index 1, with a $2.7\%$ increase in TOP$_{lsls}$. Comparing the results from index 2 and index 3, it can be seen that due to the inaccuracies and instability in the outputs of the lane segment decoder, interacting with the topology sequence decoder too early results in worse performance than using randomly ordered GT key points as key point prompts. However, compared to the results in index 1, this approach still leads to a slight improvement in topology reasoning. From the results in index 2 and index 4, it can be seen that when the predicted key points from the lane segment decoder are introduced into the key point prompts for an additional 24 epochs of interaction training between the two decoders, the mAP improves by $2.8\%$, and the TOP increases by $2.1\%$. This result indicates that sequence learning can further enhance the extraction of BEV features in the regions of interest for the lane segment decoder, specifically strengthening long-range perception and topology reasoning.

\subsection{Qualitative Results}
As shown in Fig. \ref{fig:Qualitative_results}, we visualize the lane graphs generated by LaneSegNet and Topo2Seq. In comparison, Topo2Seq generates higher-quality lane graphs with aligned endpoints, more reliable long-range perception, and accurate topological relationships. This is attributed to the advantages inherited from the interaction with the sequence topology.

\section{Conclusions}
We present Topo2Seq, a topology reasoning method enhanced by topology sequence learning. Drawing inspiration from language models, we address the limitations in long-range perception and relationship modeling inherent in DETR-based topology reasoning frameworks through sequence-to-sequence learning. By incorporating randomized order prompt-to-sequence learning, we enhance the interaction between the topology sequence decoder and the lane segment decoder. This approach enables Topo2Seq to generate lane graphs with more accurately aligned endpoints and precise topology. Experimental results on the OpenLane-V2 dataset show that Topo2Seq achieves state-of-the-art performance in topology reasoning.

\section{Acknowledgments}
This work was supported by the Basic Research Project
No. HZQB-KCZYZ-2021067 of Hetao Shenzhen HK S\&T Cooperation Zone, by Shenzhen
General Program No. JCYJ20220530143600001, by Shenzhen-Hong Kong Joint Funding No.
SGDX20211123112401002, by the Shenzhen Outstanding Talents Training Fund 202002, by Guangdong Research Project No. 2017ZT07X152 and No. 2019CX01X104, by the Guangdong Provincial Key Laboratory of Future Networks of Intelligence (Grant No. 2022B1212010001), by the
Guangdong Provincial Key Laboratory of Big Data Computing, CHUK-Shenzhen, by the NSFC
61931024\&12326610, by the Key Area R\&D Program of Guangdong Province with grant No.
2018B030338001, by the Shenzhen Key Laboratory of Big Data and Artificial Intelligence (Grant
No. ZDSYS201707251409055), and by Tencent \& Huawei Open Fund.

\bibliography{aaai25}

\begin{thebibliography}{31}
\providecommand{\natexlab}[1]{#1}

\bibitem[{Caesar et~al.(2020)Caesar, Bankiti, Lang, Vora, Liong, Xu, Krishnan, Pan, Baldan, and Beijbom}]{caesar2020nuscenes}
Caesar, H.; Bankiti, V.; Lang, A.~H.; Vora, S.; Liong, V.~E.; Xu, Q.; Krishnan, A.; Pan, Y.; Baldan, G.; and Beijbom, O. 2020.
\newblock nuscenes: A multimodal dataset for autonomous driving.
\newblock In \emph{Proceedings of the IEEE/CVF conference on computer vision and pattern recognition}, 11621--11631.

\bibitem[{Can et~al.(2021)Can, Liniger, Paudel, and Van~Gool}]{can2021structured}
Can, Y.~B.; Liniger, A.; Paudel, D.~P.; and Van~Gool, L. 2021.
\newblock Structured bird's-eye-view traffic scene understanding from onboard images.
\newblock In \emph{Proceedings of the IEEE/CVF International Conference on Computer Vision}, 15661--15670.

\bibitem[{Carion et~al.(2020)Carion, Massa, Synnaeve, Usunier, Kirillov, and Zagoruyko}]{carion2020end}
Carion, N.; Massa, F.; Synnaeve, G.; Usunier, N.; Kirillov, A.; and Zagoruyko, S. 2020.
\newblock End-to-end object detection with transformers.
\newblock In \emph{European conference on computer vision}, 213--229. Springer.

\bibitem[{Chen et~al.(2021)Chen, Saxena, Li, Fleet, and Hinton}]{chen2021pix2seq}
Chen, T.; Saxena, S.; Li, L.; Fleet, D.~J.; and Hinton, G. 2021.
\newblock Pix2seq: A language modeling framework for object detection.
\newblock \emph{arXiv preprint arXiv:2109.10852}.

\bibitem[{Chen et~al.(2022)Chen, Saxena, Li, Lin, Fleet, and Hinton}]{chen2022unified}
Chen, T.; Saxena, S.; Li, L.; Lin, T.-Y.; Fleet, D.~J.; and Hinton, G.~E. 2022.
\newblock A unified sequence interface for vision tasks.
\newblock \emph{Advances in Neural Information Processing Systems}, 35: 31333--31346.

\bibitem[{Deng et~al.(2009)Deng, Dong, Socher, Li, Li, and Fei-Fei}]{deng2009imagenet}
Deng, J.; Dong, W.; Socher, R.; Li, L.-J.; Li, K.; and Fei-Fei, L. 2009.
\newblock Imagenet: A large-scale hierarchical image database.
\newblock In \emph{2009 IEEE conference on computer vision and pattern recognition}, 248--255. Ieee.

\bibitem[{Devlin et~al.(2018)Devlin, Chang, Lee, and Toutanova}]{devlin2018bert}
Devlin, J.; Chang, M.-W.; Lee, K.; and Toutanova, K. 2018.
\newblock BERT: Pre-training of Deep Bidirectional Transformers for Language Understanding.
\newblock \emph{arXiv preprint arXiv:1810.04805}.

\bibitem[{He et~al.(2016)He, Zhang, Ren, and Sun}]{he2016deep}
He, K.; Zhang, X.; Ren, S.; and Sun, J. 2016.
\newblock Deep residual learning for image recognition.
\newblock In \emph{Proceedings of the IEEE conference on computer vision and pattern recognition}, 770--778.

\bibitem[{Kingma and Ba(2015)}]{diederik2015adam}
Kingma, D.~P.; and Ba, J. 2015.
\newblock Adam: A Method for Stochastic Optimization.
\newblock In \emph{ICLR}.

\bibitem[{Li et~al.(2024)Li, Huang, Wang, Rong, Wang, and Liu}]{li2024enhancing}
Li, H.; Huang, Z.; Wang, Z.; Rong, W.; Wang, N.; and Liu, S. 2024.
\newblock Enhancing 3D Lane Detection and Topology Reasoning with 2D Lane Priors.
\newblock \emph{arXiv preprint arXiv:2406.03105}.

\bibitem[{Li et~al.(2022{\natexlab{a}})Li, Wang, Wang, and Zhao}]{li2022hdmapnet}
Li, Q.; Wang, Y.; Wang, Y.; and Zhao, H. 2022{\natexlab{a}}.
\newblock Hdmapnet: An online hd map construction and evaluation framework.
\newblock In \emph{2022 International Conference on Robotics and Automation (ICRA)}, 4628--4634. IEEE.

\bibitem[{Li et~al.(2023{\natexlab{a}})Li, Chen, Wang, Li, Yang, Geng, Jiang, Wang, Xu, Xu et~al.}]{li2023graph}
Li, T.; Chen, L.; Wang, H.; Li, Y.; Yang, J.; Geng, X.; Jiang, S.; Wang, Y.; Xu, H.; Xu, C.; et~al. 2023{\natexlab{a}}.
\newblock Graph-based topology reasoning for driving scenes.
\newblock \emph{arXiv preprint arXiv:2304.05277}.

\bibitem[{Li et~al.(2023{\natexlab{b}})Li, Jia, Wang, Chen, Jiang, Yan, and Li}]{li2023lanesegnet}
Li, T.; Jia, P.; Wang, B.; Chen, L.; Jiang, K.; Yan, J.; and Li, H. 2023{\natexlab{b}}.
\newblock Lanesegnet: Map learning with lane segment perception for autonomous driving.
\newblock \emph{arXiv preprint arXiv:2312.16108}.

\bibitem[{Li et~al.(2022{\natexlab{b}})Li, Wang, Li, Xie, Sima, Lu, Qiao, and Dai}]{li2022bevformer}
Li, Z.; Wang, W.; Li, H.; Xie, E.; Sima, C.; Lu, T.; Qiao, Y.; and Dai, J. 2022{\natexlab{b}}.
\newblock Bevformer: Learning bird’s-eye-view representation from multi-camera images via spatiotemporal transformers.
\newblock In \emph{European conference on computer vision}, 1--18. Springer.

\bibitem[{Liao et~al.(2023{\natexlab{a}})Liao, Chen, Jiang, Cheng, Zhang, Liu, Huang, and Wang}]{liao2023lane}
Liao, B.; Chen, S.; Jiang, B.; Cheng, T.; Zhang, Q.; Liu, W.; Huang, C.; and Wang, X. 2023{\natexlab{a}}.
\newblock Lane graph as path: Continuity-preserving path-wise modeling for online lane graph construction.
\newblock \emph{arXiv preprint arXiv:2303.08815}.

\bibitem[{Liao et~al.(2022)Liao, Chen, Wang, Cheng, Zhang, Liu, and Huang}]{liao2022maptr}
Liao, B.; Chen, S.; Wang, X.; Cheng, T.; Zhang, Q.; Liu, W.; and Huang, C. 2022.
\newblock Maptr: Structured modeling and learning for online vectorized hd map construction.
\newblock \emph{arXiv preprint arXiv:2208.14437}.

\bibitem[{Liao et~al.(2023{\natexlab{b}})Liao, Chen, Zhang, Jiang, Zhang, Liu, Huang, and Wang}]{liao2023maptrv2}
Liao, B.; Chen, S.; Zhang, Y.; Jiang, B.; Zhang, Q.; Liu, W.; Huang, C.; and Wang, X. 2023{\natexlab{b}}.
\newblock Maptrv2: An end-to-end framework for online vectorized hd map construction.
\newblock \emph{arXiv preprint arXiv:2308.05736}.

\bibitem[{Lin et~al.(2017{\natexlab{a}})Lin, Doll{\'a}r, Girshick, He, Hariharan, and Belongie}]{lin2017feature}
Lin, T.-Y.; Doll{\'a}r, P.; Girshick, R.; He, K.; Hariharan, B.; and Belongie, S. 2017{\natexlab{a}}.
\newblock Feature pyramid networks for object detection.
\newblock In \emph{Proceedings of the IEEE conference on computer vision and pattern recognition}, 2117--2125.

\bibitem[{Lin et~al.(2017{\natexlab{b}})Lin, Goyal, Girshick, He, and Doll{\'a}r}]{lin2017focal}
Lin, T.-Y.; Goyal, P.; Girshick, R.; He, K.; and Doll{\'a}r, P. 2017{\natexlab{b}}.
\newblock Focal loss for dense object detection.
\newblock In \emph{Proceedings of the IEEE international conference on computer vision}, 2980--2988.

\bibitem[{Liu et~al.(2023{\natexlab{a}})Liu, Yan, Jia, Li, Gao, Wang, and Zhang}]{liu2023petrv2}
Liu, Y.; Yan, J.; Jia, F.; Li, S.; Gao, A.; Wang, T.; and Zhang, X. 2023{\natexlab{a}}.
\newblock Petrv2: A unified framework for 3d perception from multi-camera images.
\newblock In \emph{Proceedings of the IEEE/CVF International Conference on Computer Vision}, 3262--3272.

\bibitem[{Liu et~al.(2023{\natexlab{b}})Liu, Yuan, Wang, Wang, and Zhao}]{liu2023vectormapnet}
Liu, Y.; Yuan, T.; Wang, Y.; Wang, Y.; and Zhao, H. 2023{\natexlab{b}}.
\newblock Vectormapnet: End-to-end vectorized hd map learning.
\newblock In \emph{International Conference on Machine Learning}, 22352--22369. PMLR.

\bibitem[{Liu et~al.(2023{\natexlab{c}})Liu, Chen, Guo, Wang, Cheng, Zhu, Zhang, Liu, and Zhang}]{liu2023vision}
Liu, Z.; Chen, S.; Guo, X.; Wang, X.; Cheng, T.; Zhu, H.; Zhang, Q.; Liu, W.; and Zhang, Y. 2023{\natexlab{c}}.
\newblock Vision-based uneven bev representation learning with polar rasterization and surface estimation.
\newblock In \emph{Conference on Robot Learning}, 437--446. PMLR.

\bibitem[{Liu et~al.(2024)Liu, Zhang, Liu, Zhao, and Xu}]{liu2024leveraging}
Liu, Z.; Zhang, X.; Liu, G.; Zhao, J.; and Xu, N. 2024.
\newblock Leveraging Enhanced Queries of Point Sets for Vectorized Map Construction.
\newblock \emph{arXiv preprint arXiv:2402.17430}.

\bibitem[{Lu et~al.(2023)Lu, Peng, Cai, Xu, Li, Wen, Zhang, and Zhang}]{lu2023translating}
Lu, J.; Peng, R.; Cai, X.; Xu, H.; Li, H.; Wen, F.; Zhang, W.; and Zhang, L. 2023.
\newblock Translating Images to Road Network: A Non-Autoregressive Sequence-to-Sequence Approach.
\newblock In \emph{Proceedings of the IEEE/CVF International Conference on Computer Vision}, 23--33.

\bibitem[{Ma et~al.(2024)Ma, Liang, Wen, Lu, and Wan}]{ma2024roadpainter}
Ma, Z.; Liang, S.; Wen, Y.; Lu, W.; and Wan, G. 2024.
\newblock RoadPainter: Points Are Ideal Navigators for Topology transformER.
\newblock \emph{arXiv preprint arXiv:2407.15349}.

\bibitem[{Peng et~al.(2024)Peng, Cai, Xu, Lu, Wen, Zhang, and Zhang}]{peng2024lanegraph2seq}
Peng, R.; Cai, X.; Xu, H.; Lu, J.; Wen, F.; Zhang, W.; and Zhang, L. 2024.
\newblock LaneGraph2Seq: Lane Topology Extraction with Language Model via Vertex-Edge Encoding and Connectivity Enhancement.
\newblock In \emph{Proceedings of the AAAI Conference on Artificial Intelligence}, volume~38, 4497--4505.

\bibitem[{Vaswani(2017)}]{vaswani2017attention}
Vaswani, A. 2017.
\newblock Attention is all you need.
\newblock \emph{arXiv preprint arXiv:1706.03762}.

\bibitem[{Wang et~al.(2024)Wang, Li, Li, Chen, Sima, Liu, Wang, Jia, Wang, Jiang et~al.}]{wang2024openlane}
Wang, H.; Li, T.; Li, Y.; Chen, L.; Sima, C.; Liu, Z.; Wang, B.; Jia, P.; Wang, Y.; Jiang, S.; et~al. 2024.
\newblock Openlane-v2: A topology reasoning benchmark for unified 3d hd mapping.
\newblock \emph{Advances in Neural Information Processing Systems}, 36.

\bibitem[{Wilson et~al.(2023)Wilson, Qi, Agarwal, Lambert, Singh, Khandelwal, Pan, Kumar, Hartnett, Pontes et~al.}]{wilson2023argoverse}
Wilson, B.; Qi, W.; Agarwal, T.; Lambert, J.; Singh, J.; Khandelwal, S.; Pan, B.; Kumar, R.; Hartnett, A.; Pontes, J.~K.; et~al. 2023.
\newblock Argoverse 2: Next generation datasets for self-driving perception and forecasting.
\newblock \emph{arXiv preprint arXiv:2301.00493}.

\bibitem[{Wu et~al.(2023)Wu, Chang, Jia, Liu, Wang, and Shen}]{wu2023topomlp}
Wu, D.; Chang, J.; Jia, F.; Liu, Y.; Wang, T.; and Shen, J. 2023.
\newblock Topomlp: An simple yet strong pipeline for driving topology reasoning.
\newblock \emph{arXiv preprint arXiv:2310.06753}.

\bibitem[{Zhu et~al.(2020)Zhu, Su, Lu, Li, Wang, and Dai}]{zhu2020deformable}
Zhu, X.; Su, W.; Lu, L.; Li, B.; Wang, X.; and Dai, J. 2020.
\newblock Deformable detr: Deformable transformers for end-to-end object detection.
\newblock \emph{arXiv preprint arXiv:2010.04159}.

\end{thebibliography}

\end{document}